\documentclass[11pt,a4paper]{article}
\usepackage[hyperref]{emnlp2020}
\usepackage{times}
\usepackage{latexsym}
\usepackage{caption}
\usepackage{subcaption}
\usepackage{color}
\usepackage{mathtools}
\usepackage{amsmath}
\usepackage{amsfonts}
\usepackage{amsthm}
\newtheorem{definition}{Definition}
\newtheorem{proposition}{Proposition}

\usepackage{microtype}
\usepackage{enumitem}
\usepackage{pgfplots}
\usepackage{multirow}
\usepackage{rotating}

\aclfinalcopy 
\def\aclpaperid{10} 

\newcommand{\specialcell}[2][c]{%
\begin{tabular}[#1]{@{}c@{}}#2\end{tabular}}

\title{Information Extraction from Swedish Medical Prescriptions with Sig-Transformer Encoder}

\author{John Pougu\'e Biyong$^1$ \quad Bo Wang$^2$$^,$$^3$ \quad Terry Lyons$^1$$^,$$^3$ \quad Alejo J Nevado-Holgado$^2$ \\
  $^1$Mathematical Institute, University of Oxford, UK\\
  $^2$Department of Psychiatry, University of Oxford, UK\\
  $^3$The Alan Turing Institute, London, UK \\
\texttt{john.pougue-biyong@maths.ox.ac.uk, bo.wang@psych.ox.ac.uk}}

\date{08/08/2020}

\begin{document}

\maketitle

\begin{abstract}
Relying on large pretrained language models such as Bidirectional Encoder Representations from Transformers (BERT) for encoding and adding a simple prediction layer has led to impressive performance in many clinical natural language processing (NLP) tasks. In this work, we present a novel extension to the Transformer architecture, by incorporating signature transform with the self-attention model. This architecture is added between embedding and prediction layers. Experiments on a new Swedish prescription data show the proposed architecture to be superior in two of the three information extraction tasks, comparing to baseline models. Finally, we evaluate two different embedding approaches between applying Multilingual BERT and translating the Swedish text to English then encode with a BERT model pretrained on clinical notes.
\end{abstract}


\section{Introduction}
\label{sec:intro}
Medical prescription notes written by clinicians about patients contains valuable information that the structured part of electronic health records (EHRs) does not have. Manually extracting and annotating such information from large amount of textual fields in medical records such as prescription notes is time-consuming, and expensive since it has to be provided by domain specialists who are in high demand. Information extraction (IE), a specialised area in natural language processing (NLP), refers to the automatic extraction of structured information such as concepts, entities and events from free text \cite{wang2018clinical}. While rule-base IE systems are still widely used in healthcare \cite{banda2018advances,davenport2019potential}, machine learning-based data-driven approaches have gained much more interest due to their ability to scale and learn to recognise complex patterns. In this paper, we present a set of medical prescriptions written in Swedish and develop a neural network-based system that extracts important information from the textual data.

Recent advances in pretraining large-scale contextualised language representations, including ELMo \cite{peters2018deep} and BERT \cite{devlin2018bert}, have proven to be a successful strategy for transfer learning and pushed the performance in many NLP tasks of general purpose. Clinical notes differ substantially to general-domain text and biomedical literature, in terms of its linguistic characteristics. Several studies have fine-tuned general-domain language models such as BERT on in-domain clinical text (e.g. electronic health records) \cite{si2019enhancing,peng2019transfer,alsentzer2019publicly,huang2019clinicalbert} for downstream clinical NLP tasks. As summarised in \cite{gu2020domain} many such tasks can be formulated as a classification or regression task, wherein either a simple linear layer or sequential models such as LSTM or conditional random field (CRF) are added after the BERT encoding of the input text as task-specific prediction layer. 

BERT's model architecture is based on the Transformer model \cite{vaswani2017attention}, which employs the self-attention mechanism to attend to different positions of the input sequence. As it doesn't contain any recurrence or convolution, the model has to add positional encoding in order to model the order of the sequence (e.g. word order). Signature transform, initially introduced in rough path theory as a branch of stochastic analysis, is a non-parametric approach of encoding sequential data while capturing the order information in the data. It has shown successes in the recent years in a range of machine learning tasks involving sequential modelling \cite{arribas2018signature,morrill2020utilization,tothbayesian,wang2020learning}. We hypothesise the signature transform method can be integrated in the Transformer model, in which its ability to naturally capture sequential ordering complements the ability of attending to important positions and learning long-range dependencies in self-attention. 

As a preliminary study, our aim is to develop a machine learning-based system that extracts relevant information from the Swedish medical prescription notes\footnote{Our ultimate goal is to apply such model to a much larger database. However, this is beyond the scope of this paper.}, namely \textit{quantity}, \textit{quantity tag} and \textit{indication}. The contributions of this work are as follows:
\begin{enumerate}
 \item We experiment two different approaches of embedding the Swedish prescription notes. One encodes the Swedish text directly using Multilingual BERT (M-BERT) \cite{devlin2018bert} while for the other approach we translate the prescriptions and then apply ClinicalBERT \cite{huang2019clinicalbert} that is pretrained on clinical text in English.
 \item We propose an extension to the Transformer model, named Sig-Transformer Encoder (STE), which integrates signature transform into the Transformer architecture so the order information in the prescription notes can be learnt in a more effective way. To the best of our knowledge, this is the first attempt to integrate signature transform in the Transformer architecture.
 \item We demonstrate good performance in two of the three tasks, namely \textit{quantity} and \textit{quantity tag}. As for the \textit{indication} task, we provide an analysis on why the models fail.
 \item We show one of our proposed STE-based approaches, namely M-BERT+STE, outperforming other baseline models.
\end{enumerate}

\section{Background}
\label{sec:related}
Clinical language representation has attracted an increasing interest in the natural language processing (NLP) community \cite{kalyan2020secnlp}. More recently with the advent of Bidirectional Encoder Representations from Transformers (BERT) \cite{devlin2018bert}, fine-tuning of general-domain language models has been widely adopted for many clinical NLP tasks\footnote{\citet{gu2020domain} have challenged this strategy and reported better performance for PubMed-based biomedical tasks by conducting domain-specific pretraining from scratch. However, it did not address the clinical domain and has left it for future work.}. Among them two clinicalBERT studies \cite{alsentzer2019publicly,huang2019clinicalbert} have conducted fine-tuning from either BERT or BioBERT \cite{lee2020biobert} using de-identified clinical notes from MIMIC-III \cite{johnson2016mimic}. In particular, \citet{huang2019clinicalbert} showed the effectiveness of its ClinicalBERT-based model for predicting hospital readmissions. 

Many NLP applications in the medical domain can be formulated as either token classification, sequence classification or sequence regression, in which a pretrained language model such as clinicalBERT can be used to encode the input token sequence (which returns either the encoding for every input token or just the [CLS] token) and then a task-specific prediction model is added on top to generate the final output \cite{gu2020domain}. In this section, we describe the Transformer encoder and the signature transform, which are the bases of our proposed architecture. 

\subsection{Transformer Encoder}
The Transformer model employs an encoder-decoder structure \cite{vaswani2017attention}. Its encoder, which is the model architecture of BERT, is composed of a stack of $N$ identical layers. Each layer has a multi-head self-attention mechanism followed by a position-wise fully connected feed-forward network. The attention function used in the Transformer model takes three input vectors: query ($Q$), key ($K$) and value ($V$). It generates an output vector by computing the weighted sum of the values. The weights are computed by the dot products of the query and all keys, scaled and applied a softmax function. 
\begin{equation}
\label{eq:att}
\begin{aligned}      
   \mathrm{Attention}(Q, K, V) = \mathrm{softmax}(\frac{QK^T}{\sqrt{d_k}})V  
\end{aligned}
\end{equation}
Multi-head attention splits $Q$, $K$ and $V$ into multiple heads by linearly projection, which allows the model to jointly attend to information at different positions from different representation subspaces. Each projected head goes through the scaled dot-product attention function, then concatenated and projected again to output the final values. The Transformer encoder does not explicitly model position information in its structure, and instead it requires adding representations of absolute positions (i.e. positional encoding) to its inputs. \citet{shaw2018self} presented an extension to self-attention, which incorporates relative position information for the input data. In this work, we propose to integrate signature transform with self-attention given that path signatures have been proven to be an effective way of capturing the sequential order information in the data. 

\subsection{Signature Transform}
The theory of rough paths, developed by \citet{lyons1998differential}, can be thought of as a non-linear extension of the classical theory of controlled differential equations. \textit{Path signature} or simply \textit{signature}, is an infinite collection of statistics characterising the underlying path (a discretised version of a continuous path), and \textit{signature transform} is the map from a path to its signature\footnote{We refer the reader to \cite{lyons2014rough} for a rigorous introduction of signature transform, and \cite{chevyrev2016primer} for a primer on its use in machine learning.}. Consider a $d$-dimensional time-dependent path $P$ over the time interval $[0,T]\subset \mathbb{R}$, to a continuous map $P:[0,T] \rightarrow \mathbb{R}^{d}$. The signature $S(P)$ of this path $P$ over time interval $[0,T]$ is the infinite collection of all iterated integrals of $P$ such that every continuous function of the path may be approximated arbitrarily well by a linear function of its signature:
\begin{equation*}
\begin{aligned}
    S(P)_{0,T} = (1, S(P)_{0,T}^1, \dots, S(P)_{0,T}^d, S(P)_{0,T}^{1,1}, \dots)
\end{aligned}
\end{equation*}
\noindent where the $0^{th}$ term is 1 by convention, and the superscripts of the terms after the $0^{th}$ term run along the set of all multi-index $\left \{ (i_1, \dots, i_k)|k\geq 1, i_1, \dots, i_k \in [d] \right \}$ with the coordinate iterated integral being:
\begin{equation*}
    S(P)_{0,T}^{i_1, \dots, i_k} = \underset{\underset{t_{1}, \dots, t_{k} \in [0, T]}{t_{1} < \dots < t_{k}}} { \int \dots \int} dP_{t_{1}}^{i_1} \otimes \dots \otimes dP_{t_{k}}^{i_k}
\end{equation*}
\noindent where $\forall k \geq 1$, $P_t \in R^d$, $\forall t \in [0, T]$. $S(P)_{0,T}^{i_1, \dots, i_k}$ is termed as the $k$th level of the signature. In practice we truncate the signature to order $n$, where the degree of its iterated integrals is no greater than $n$. This ensures the path signature has finite dimensional representation. Let $TS(P)_{0,T}^n$ denote the truncated signature of $P$ of order $n$, i.e.
\begin{equation*}
    TS(P)_{0,T}^n = (1, S(P)_{0,T}^1, \dots, S(P)_{0,T}^{k_{n}})
\end{equation*}
\noindent Therefore the dimensionality of the truncated path signature is $(d^{n+1}-d)(d-1)^{-1}$. We describe signature transform in more (mathematical) details in Appendix~\ref{sec:bg-sig}.

In recent years using path signatures as features in a suitable neural network model has shown success in various applications, such as online handwritten Chinese character recognition \cite{yang2016deepwriterid,xie2018learning}, action recognition in videos \cite{yang2017leveraging} and speech emotion recognition \cite{wang2019path}. More recently, \citet{kidger2019deep} proposed to use signature transform deeper within a network, rather than as a feature transformation. \citet{signatory} developed differentiable computation of signature transform on both CPU and GPU. To the best of our knowledge, we are the first to incorporate signature transform in the Transformer architecture.

\begin{figure}[htb]
 \begin{center}
  \includegraphics[width=0.4\textwidth]{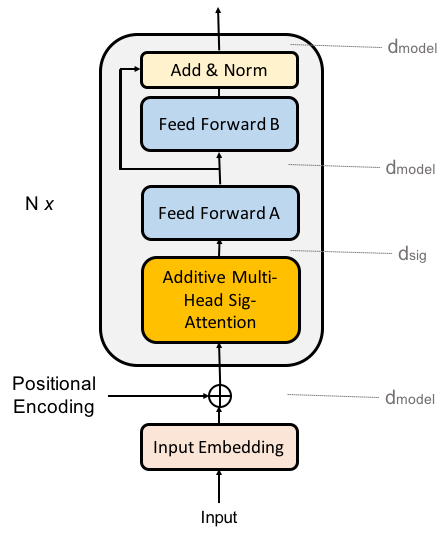}
  \caption{The Sig-Transformer Encoder. $d_{sig}$ is the dimension size in the truncated signature, and $d_{model}$ is the dimension size in the embedding layer as well as before and after the second feed-forward sub-layer.}
  \label{fig:model-arch}
 \end{center}
\end{figure}

\begin{figure*}[h]
 \begin{center}
  \includegraphics[width=0.85\textwidth]{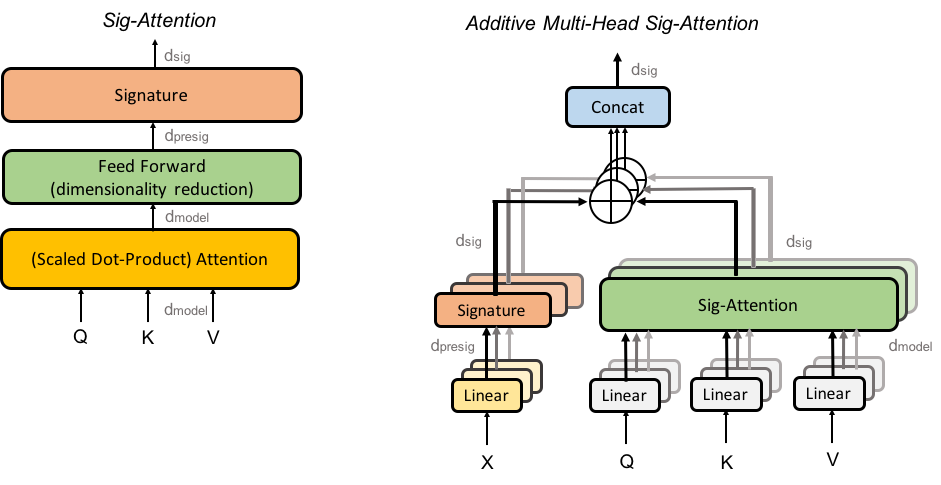}
  \caption{(left) Sig-Attention mechanism; (right) Additive Multi-Head Sig-Attention.}
  \label{fig:sig-att}
 \end{center}
\end{figure*}

\section{Sig-Transformer Encoder}
\label{sec:model}
In this section, we present our proposed extension to the Transformer Encoder (TE), named Sig-Transformer Encoder (STE).
As depicted in Figure~\ref{fig:model-arch}, STE consists of a stack of $N$ identical layers, and each layer has three sub-layers. It replaces the multi-head self-attention sub-layer in TE with our Additive Multi-Head Sig-Attention mechanism, followed by two position-wise fully connected feed-forward sub-layers. Same as the Transformer model, we also employ a residual connection \cite{he2016deep} at the last sub-layer. followed by layer normalisation \cite{ba2016layer}. The two feed-forward sub-layers and the embedding layers, produce output of the same dimension $d_{model} = 768$.

\subsection{Sig-Attention}
The core idea behind our proposed Additive Multi-Head Sig-Attention is the Sig-Attention mechanism (Figure~\ref{fig:sig-att} left). It takes the query ($Q$), key ($K$) and value ($V$) vectors as input, and performs three consecutive transformations: (1), scaled dot-product attention; (2), position-wise fully linear map, for dimensionality reduction; (3), signature transform. Its output is computed as follows:
\begin{equation*}
\label{eq:sig-att}
\begin{aligned}      
 &  \mathrm{SigAttention}(Q, K, V) =  \\
 &  \mathrm{ReducedSig}(\mathrm{Attention}(Q, K, V))
\end{aligned}
\end{equation*}
\noindent where
\begin{itemize}
    \item $\mathrm{Attention}(Q, K, V)$ is as defined in Equation~\eqref{eq:att};
    \item $\mathrm{ReducedSig}(x) = S^{N}(xW + b)$ for any $x$;
    \item $S^N$ is the signature truncated at a given order $N$;
    \item $W$ and $b$ are respectively the weight and bias matrices of the linear map.
\end{itemize}
As the size of the truncated signature increases exponentially with its input dimension, it is important to reduce the size of attended context vector before applying signature transform. We refer the readers to Table~\ref{tab:dims} for examples of how the size of the truncated signature ($d_{sig}$) is determined by its input dimension ($d_{presig}$) and the order of signature truncation ($order_{sig}$).

\subsection{Additive Multi-Head Sig-Attention}
In the proposed Additive Multi-Head Sig-Attention model, we combine the information from the input sequence with the output of Sig-Attention, in different representation subspaces (Figure~\ref{fig:sig-att} right). The model takes the embedding of an input sequence $X$ as well as the queries, keys and values as input:
\begin{equation*}  
\label{eq:amhsa}
\begin{aligned} 
& \mathrm{AdditiveMultiHead}(X, Q, K, V) =\\
&\qquad \qquad \qquad \qquad \mathrm{Concat}(\mathrm{head_1}, ..., \mathrm{head_h})\\     
&\text{where}~\mathrm{head_i} = \mathrm{ReducedSig}(X) + \\
&\qquad \mathrm{SigAttention}(QW^Q_i, KW^K_i, VW^V_i) 
\end{aligned}
\end{equation*}  
\noindent and $h$ is the number of parallel heads, the projections are parameter matrices $W^Q_i \in \mathbb{R}^{d_{\text{model}} \times d_k}$, $W^K_i \in \mathbb{R}^{d_{\text{model}} \times d_k}$, $W^V_i \in \mathbb{R}^{d_{\text{model}} \times d_k}$ and $d_k = d_{model}/h$.

As shown in Figure~\ref{fig:sig-att} (right), we linearly project the queries, keys and values as well as the input embeddings to $d_{model}$ and $d_{presig}$ dimensions. Then the query, key and value vectors will be fed into the Sig-Attention mechanism, while having been reduced its dimensions by the linear projection the input vector $X$ will be applied to signature transform. The two output signatures are added, resulting in one signature vector $s_{i} \in \mathbb{R}^{d_{sig}}$ for each head $i$. Finally all the signature vectors are concatenated and passed onto the next sub-layer.

\subsection{Position-wise Feed-Forward Networks}
The second sub-layer in Figure~\ref{fig:model-arch} is a linear transformation $W \in \mathbb{R}^{d_{sig} \times d_{model}}$ which is applied to each position separately and identically. It simply brings back the dimensions from $d_{sig}$ to $d_{model}$. The third sub-layer is a ReLU activation followed by another linear transformation. The dimensionality of its input and output is $d_{model}$.

\subsection{Dropout}
Dropout is applied to the output of the three sub-layers as well as after the ReLU activation in the third sub-layer. We use a rate of $P_{drop}=0.1$, as is used in \cite{vaswani2017attention}. Moreover, residual connection and layer normalisation are used only in the third sub-layer.

\section{Data}
\label{sec:data}
Our proposed Sig-Transformer Encoder can be used for general purpose representation learning, and any classifier or regressor can be added on top of it to perform down-stream tasks. We conduct all of our experiments with a medical prescription data for the class of ``beta-blockers (ATC code C07)'', provided by the Karolinska University Hospital. It is sampled from a much larger database of 41 million medical prescription records. This dataset contains 3,852 non-duplicated prescriptions written in Swedish. Some examples of annotated prescriptions are shown in Table~\ref{tab:data-example}.


\begin{table*}[htb]
\begin{center}
\scalebox{0.85}{
\begin{tabular}{|c|c|c|c|c|}
\hline
\textbf{Swedish} & \textbf{Translated English} & \textbf{Indication} & \textbf{Quantity} & \textbf{Quantity Tag}  \\
\hline
\specialcell{1 TABLETT FOREBYGGANDE \\ MOT MIGRAN} & \specialcell{1 tablet prevention \\ against migrain} & Migraine & 1 & Standard\\
 \hline
\specialcell{MOT HOGT BLODTRYCK \\ OCHHJARTKLAPPNING \\ 2 TABLETTER KLOCKGAN 08:00 \\ 1 TABLETT KLOCKAN 18:00.} & \specialcell{Against high blood pressure \\ and heart palpitations \\ 2 tablets at 08:00, \\ 1 tablet at 18:00.} & \specialcell{Cardiac-\\hypertension} & 3 & Standard  \\
 \hline
\specialcell{2 TABLETTER KL. 08,\\ 2 TABLETT KL. 20. DAGLIGEN.\\ OBS KVALLSDOSEN HAR\\ HOJTS JFRT MED 070427.} & \specialcell{2 tablets kl. 08,\\ 2 tablet kl. 20. Daily.\\ Note the evening box has\\ hojts jfrt with 070427.} & \specialcell{NA} & 4 & Standard \\
\hline
\specialcell{1 tablett vid behov mot stress} & \specialcell{1 tablet if needed against stress} & Anxiety & 1 & PRN \\
 \hline
\specialcell{FOR BLODTRYCK \\ OCH HJARTRYTM -} & \specialcell{For blood pressure \\ and heart rhythm -} & \specialcell{Cardiac-\\hypertension-\\dysrhythmia} & 0 & Not Specified \\
 \hline
\specialcell{1 tabl pA morgonen och \\ en halv tabl pA kvAllen \\ fOr hjArtrytmen.} & \specialcell{1 table in the morning and \\ a half table in the evening \\ for the heart rhythem.} & \specialcell{Cardiac-\\dysrhythmia} & 1.5 & Standard \\
 \hline
\specialcell{1-2 tablett 2 gAnger dagligen} & \specialcell{1-2 tablets 2 times daily} & NA & 3 & Complex \\
 \hline
\end{tabular}}
\end{center}
\caption{Example prescriptions with translations and annotations. The second column is the English translation obtained from Google Translate API. The last three columns are the labels of interest which we aim to extract automatically. Some longer example prescriptions can be viewed in Table~\ref{tab:data-example-long}.}
\label{tab:data-example}
\end{table*}

The medical practitioners at the hospital have annotated three labels, in which we use \textsc{Quantity} for regression while \textsc{Quantity tag} and \textsc{Indication} as two classification tasks:
\begin{itemize}[noitemsep,topsep=5pt,parsep=3pt,partopsep=3pt]
  \item \textsc{Quantity}: the total amount of tablets or capsules prescribed. The values are multiples of 0.5;
  \item \textsc{Quantity tag} (5 classes): the label or tag of the quantity prescribed to the patient;
  \begin{enumerate}
    \item \textit{Not Specified}: the quantity was not specified in the prescription;
    \item \textit{Complex} : a range of quantities was given. In that case, the quantity is an average between the minimum and the maximum quantities;
    \item \textit{PRN} : prescription to take only if needed;
    \item \textit{As Per Previous Prescription}: refers to guidance in previous prescription;
    \item \textit{Standard}: standard prescription;
  \end{enumerate}
  \item \textsc{Indication} (5 classes): the purpose of the prescription. It originally had 44 classes where many account for only one record\footnote{The original 44 classes and the number of prescriptions per class, can be found in Figure~\ref{fig:indication-original}.}. The medical practitioners have aggregated them into 5 medically meaningful classes: \textit{Cardiac}, \textit{Tremors}, \textit{Migraine}, \textit{Others}, and \textit{NA} (Not Annotated). 
\end{itemize}

One challenge of the data is the remarkable class imbalance in its labels, as shown in Table~\ref{tab:class-dist}. It's still prominent in the \textsc{Indication} field after aggregating the original classes, where \textit{Cardiac} and \textit{NA} account for majority of the instances, while the other 3 classes are seldom annotated in the data. Another challenge, which is prevalent in electronic health record (EHR) datasets, is the free-form nature of its text and the writing styles vary considerably between different doctors. 

\begin{table}[!htbp]
\centering
\scalebox{0.85}{
\begin{tabular}{ |c|c|c|c|c|c|  }
 \hline
 \multicolumn{3}{|c|}{\textsc{Quantity Tag}} &
 \multicolumn{3}{|c|}{\textsc{Indication}} \\
 \hline
 Standard & 3489 & 90.6\% & Cardiac & 2980 & 77.4\% \\
 PRN & 89 & 2.3\% & Tremors & 81 & 2.1\% \\
 APPP & 40 & 1.0\% & Migraine & 69 & 1.8\% \\
 Complex & 29 & 0.8\% & Other & 15 & 0.4\% \\
 NS & 205 & 5.3\% & NA & 707 & 18.4\% \\
 \hline
\end{tabular}
}
\caption{Class distribution for \textsc{Quantity Tag} and \textsc{Indication}. APPP is an abbreviation for \textit{As Per Previous Prescription}, NS stands for \textit{Not Specified} and NA stands for \textit{Not Annotated}.}
\label{tab:class-dist}
\end{table}

\begin{figure}[htb]
 \begin{center}
  \includegraphics[width=0.5\textwidth]{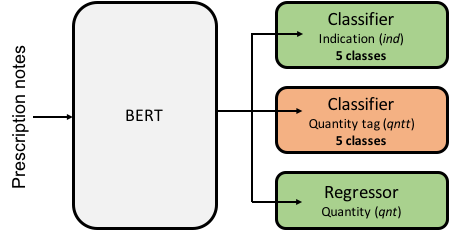}
  \caption{Multi-task learning architecture without Sig-Transformer Encoder (STE).}
  \label{fig:mtask-1}
 \end{center}
\end{figure}

\begin{figure}[htb]
 \begin{center}
  \includegraphics[width=0.5\textwidth]{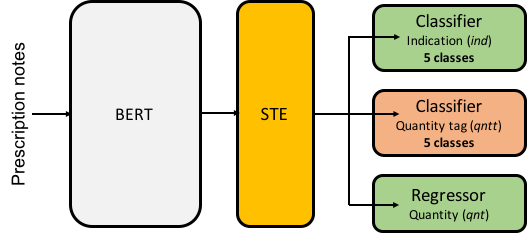}
  \caption{Our proposed multi-task learning architecture with Sig-Transformer Encoder (STE).}
  \label{fig:mtask-2}
 \end{center}
\end{figure}

\subsection{Data Preprocessing}
\label{sec:preprop}
As mentioned in Section~\ref{sec:intro} we experiment two different embedding models for the Swedish prescription records, in which one of them, namely ClinicalBERT \cite{huang2019clinicalbert}, requires its input text to be English. We use the Google Translate API to obtain translated English prescriptions. Minimal text preprocessing was taken, and all prescription text including non-ASCII characters (e.g. \aa, \"a, \"o) was read with UTF-8 encoding.

\section{Multi-Task Learning}
\label{sec:mtl}
Our objective is to automatically find the \textsc{Quantity} of prescribed medicine as well as \textsc{Quantity tag} and \textsc{Indication}
of the prescription, for each prescription note. We formulate this as a multi-task learning problem consisting of a regression task and two classification tasks. As depicted in Figure~\ref{fig:mtask-1}, such multi-task learning model uses BERT or any of its variants to encode each input prescription note represented by the respective [CLS] token, then we can add three separate predictors on top of it and all three jointly learn with the embedding layer. We use this architecture as one of our baseline approaches.

In our proposed multi-task learning architecture, we add Sig-Transformer Encoder (STE) in-between the embedding layer (BERT) and the three predictors in order to learn the sequential order information in the input text encoded by the token-level BERT representations (instead from [CLS]). The three predictors and STE are jointly trained. We compare the two architectures and other baseline models in Section~\ref{sec:exp}.

\section{Experiments}
\label{sec:exp}

\subsection{Experiment Setup}
As described in Section~\ref{sec:related}, BERT and its variants have been used in a range of clinical machine learning tasks. Considering our data is in Swedish we explore two different approaches for encoding the prescription notes: (1), Apply Multilingual BERT (M-BERT) \cite{devlin2018bert} directly to the Swedish text; (2), Translate the prescriptions to English as described in Section~\ref{sec:preprop}, and then encode the translated text with ClinicalBERT \cite{huang2019clinicalbert}. ClinicalBERT is fine-tuned on clinical notes, therefore it has embedded more domain knowledge and has proven to be more effective over BERT-base on several clinical tasks. However, any translation error would affect the performance of subsequent models. 

We use the PyTorch implementation of Transformers by Hugging Face\footnote{https://github.com/huggingface/transformers} for loading M-BERT and ClinicalBERT, and \textit{Signatory}\footnote{https://github.com/patrick-kidger/signatory} \cite{signatory} for differentiable computations of signatures on GPU. We use a linear layer as the classifier for \textsc{Quantity tag}, and two-layer network for \textsc{Quantity} and \textsc{Indication} respectively. More details of the implementation choices are described in Appendix~\ref{sec:imp-de}. We use cross-entropy (CE) loss for the two classification tasks and mean-square error (MSE) for the regression task. The overall loss function is a weighted average of the loss functions of all three tasks:
\begin{equation*}  
\begin{aligned} 
&\mathcal{L} = \\
&\alpha_{qnt} \times MSE_{qnt} + \beta_{qntt} \times CE_{qntt} + \beta_{ind} \times CE_{ind}
\end{aligned}
\end{equation*} 
\noindent where \textit{qnt} stands for quantity, \textit{qntt} stands for quantity tag, \textit{ind} stands for indication, and $\alpha_{qnt}+\beta_{qntt}+\beta_{ind} = 1$. $\alpha_{qnt}$, $\beta_{qntt}$ and $\beta_{ind}$ are parameters to be optimised. 

To compare model performance we use the simple multi-task learning models described in Section~\ref{sec:mtl} without adding STE (Figure~\ref{fig:mtask-1}), as the baseline systems. We also replace STE with LSTM for comparison as both learn sequential information from the input data\footnote{Unlike the Transformer model, recurrent neural networks (RNNs) such as LSTM, model relative and absolute positions along the time dimension directly through their sequential structure \cite{shaw2018self}.}. All experiments are conducted using 5-fold cross validation, where we use 3 folds for training, 1 fold for validation and 1 fold as the test set at each iteration. We repeat the experiments for each model with 10 different random seeds and average the scores as the final result.


\subsection{Results}
\label{sec:res}
The results for all three tasks are summarised in Table~\ref{tab:res}. We first notice all models have failed to recognise the \textsc{Indication} classes, which we will discuss in Section~\ref{sec:res-ind}. In this section we mainly discuss the performance for \textsc{Quantity} and \textsc{Quantity Tag}.

Among the three models without the use of LSTM or STE, both ClinicalBERT and Multilingual BERT (M-BERT) outperform the BERT-base model in all three tasks. As for the three LSTM-based models, using the BERT-base representations achieves the best performance while ClinicalBERT and M-BERT are comparable with each other. Among our proposed STE-based models, M-BERT+STE outperforms the other two variants.

Comparing the results across the board, adding LSTM or STE improves the model performance in most cases except for ClinicalBERT, where the addition of LSTM or STE has worsened the performance for ClinicalBERT. We think the potential discrepancies in the translation of the Swedish prescriptions have possibly affected the working of LSTM and STE, as both models take all the translated token embeddings of each prescription as input where the simple ClinicalBERT baseline only takes the encoding of the [CLS] token as input. Comparing between the two different prescription embedding approaches, translating them then apply ClinicalBERT clearly beats using the multilingual-BERT when LSTM or STE is not added. When we incorporate LSTM or STE in the model, the multilingual embedding approach has obtained better results in most cases. Overall, one of our proposed models, M-BERT+STE, achieves the best performance for the \textsc{Quantity} and \textsc{Quantity Tag} tasks\footnote{The results obtained by M-BERT+STE and ClinicalBERT are different at the significance level of 0.1, with $\textup{p}=0.068$ and $0.065$ for \textsc{Quantity} and \textsc{Quantity Tag} respectively, using Wilcoxon signed-rank test.}. 

In Table~\ref{tab:res-xl-quant} we present results obtained by each model across different classes of \textsc{Quantity Tag}. We can see the \textit{Standard} class (i.e. standard prescription) followed by \textit{PRN} (i.e. prescription to take only if needed) and \textit{APPP} (i.e. as per previous prescription) have received better classifications by the three STE models. The models also performed well in the \textit{Not Specified} class. The \textit{Complex} class refers to a range of quantities given in the prescription. As expected, many models do not perform well in this class. The overall best performing model, M-BERT+STE, also performs consistently well across different classes of \textsc{Quantity Tag}.

\begin{table*}[htb]
\begin{center}
\scalebox{1.0}{
\begin{tabular}{lccc}
\hline \bf Model & \bf \textsc{Quantity} & \bf \textsc{Quantity Tag} & \bf \textsc{Indication}\\ \hline
Base & 0.50 & 0.60 & 0.08 \\
ClinicalBERT & 0.21 & 0.89 & 0.09 \\
M-BERT & 0.41 & 0.63 & \textbf{0.10} \\
\hline
Base + LSTM & 0.45 & 0.76 & 0.06 \\
ClinicalBERT + LSTM & 0.47 & 0.64 & \textbf{0.10} \\
M-BERT + LSTM & 0.50 & 0.68 & 0.02 \\
\hline
Base + STE & 0.23 & 0.78 & 0.03 \\
ClinicalBERT + STE & 0.36 & 0.77 & 0.06 \\
M-BERT + STE & \textbf{0.15} & \textbf{0.92} & 0.05 \\
\hline
\end{tabular}}
\end{center}
\caption{Performance comparison between our proposed STE-based approaches and baseline models. The \textsc{Quantity} task is measured in mean squared error (MSE), while for \textsc{Quantity Tag} and \textsc{Indication} we use Macro F-1 score. Base refers the original BERT-base model, M-BERT is the Multilingual BERT model pretrained on Swedish text data. STE refers to Sig-Transformer Encoder.}
\label{tab:res}
\end{table*}

\subsection{The difficult case of Indication}
\label{sec:res-ind}
The class imbalance is remarkable for \textsc{Quantity Tag} and \textsc{Indication} in this dataset, as described in Section~\ref{sec:data}. The strong results obtained from many models for \textsc{Quantity Tag} suggests albeit the class imbalance the boundaries between its classes are distinguishable and can be learnt from fewer examples. \textsc{Indication} is a more challenging task. The aggregation of the original 44 classes to the final 5 classes has reduced the severity of the class imbalance issue\footnote{We have experimented with the original 44 classes, and obtained terrible results. We have also tried setting class weights to alleviate the class imbalance issue with no significant difference observed in the final results.}. However, the new classification becomes much less distinguishable and the new classes are less distinct as each class now contains a range of topics, as shown in Figure~\ref{fig:indication-original}. Such low scores are also observed across different classes of \textsc{Indication} in Table~\ref{tab:res-xl-ind}.

\subsection{Ablation Study}
\label{sec:res-pe}
The positional encoding proposed in \cite{vaswani2017attention} is based on sinusoids of varying frequency, and the authors hypothesise this would help the model to generalise to sequences of variable lengths during training. Our proposed architecture has both the absolute positional representations of words and the signature transform that is also invariant to the sequence length for encoding order of events. Here we conduct ablation study by removing the positional encoding or the signature transform components from the best performing model, M-BERT+STE, and compare the results. As shown in Table~\ref{tab:res-posenc}, we obtain worse performance without positional encoding or without signature transform, for all three tasks, suggesting the contributions of using the absolute positional encoding and the proposed integration of signature transform.

\begin{table}[htb]
\begin{center}
\scalebox{0.90}{
\begin{tabular}{lccc}
\hline \bf Model & \bf Quantity & \bf Quantity Tag & \bf Indication\\ \hline
Full model & 0.15 & 0.92 & 0.05 \\
w/o PE & 0.38 & 0.81 & 0.01 \\
w/o ST & 0.31 & 0.77 & 0.02 \\
\hline
\end{tabular}}
\end{center}
\caption{Ablation study of the best performing M-BERT + STE model, removing positional encoding (PE) or signature transform (ST).}
\label{tab:res-posenc}
\end{table}

\section{Conclusions and Future Work}
\label{sec:conclusions}
In this work, we propose a new extension to the Transformer architecture, named Sig-Transformer Encoder (STE), by incorporating signature transform with the self-attention mechanism. As a preliminary study, we aimed to automatically extract information related to \textit{quantity}, \textit{quantity tag} and the \textit{indication} label from a Swedish medical prescription dataset. We demonstrated good performance in two of the three tasks in a multi-task learning framework. Lastly, we compared two embedding approaches between applying ClinicalBERT (on translated text) and Multilingual BERT (M-BERT). Although one of our proposed models, namely M-BERT+STE, reported the best performance for \textit{quantity} and \textit{quantity tag}, all models failed to perform on the \textit{indication} task which we provided possible explanations. 

For future work, we plan to apply the proposed STE models to the much larger prescription database and investigate ways to improve the labelling for \textit{indication}. We also plan to benchmark our models on other clinical tasks such as predicting hospital readmission \cite{huang2019clinicalbert}, as well as the more general NLP tasks evaluated in \cite{devlin2018bert}. Moreover, a further investigation to better understand the contribution of signature transform and the interplay between signature and attentions would be very insightful. 


\section*{Acknowledgments}
This work was supported by the MRC Mental Health Data Pathfinder award to the University of Oxford [MC\_PC\_17215], by the NIHR Oxford Health Biomedical Research Centre and by the The Alan Turing Institute under the EPSRC grant EP/N510129/1. We would like to thank Dr. Remus-Giulio Anghel and Dr. Yasmina Molero Plaza for providing the data and sharing information about the data.

\bibliographystyle{acl_natbib}
\bibliography{anthology,emnlp2020}

\begin{thebibliography}{30}
\expandafter\ifx\csname natexlab\endcsname\relax\def\natexlab#1{#1}\fi

\bibitem[{Alsentzer et~al.(2019)Alsentzer, Murphy, Boag, Weng, Jindi, Naumann,
  and McDermott}]{alsentzer2019publicly}
Emily Alsentzer, John Murphy, William Boag, Wei-Hung Weng, Di~Jindi, Tristan
  Naumann, and Matthew McDermott. 2019.
\newblock Publicly available clinical bert embeddings.
\newblock In \emph{Proceedings of the 2nd Clinical Natural Language Processing
  Workshop}, pages 72--78.

\bibitem[{Arribas et~al.(2018)Arribas, Goodwin, Geddes, Lyons, and
  Saunders}]{arribas2018signature}
Imanol~Perez Arribas, Guy~M Goodwin, John~R Geddes, Terry Lyons, and Kate~EA
  Saunders. 2018.
\newblock A signature-based machine learning model for distinguishing bipolar
  disorder and borderline personality disorder.
\newblock \emph{Translational psychiatry}, 8(1):1--7.

\bibitem[{Ba et~al.(2016)Ba, Kiros, and Hinton}]{ba2016layer}
Jimmy~Lei Ba, Jamie~Ryan Kiros, and Geoffrey~E Hinton. 2016.
\newblock Layer normalization.
\newblock \emph{arXiv preprint arXiv:1607.06450}.

\bibitem[{Banda et~al.(2018)Banda, Seneviratne, Hernandez-Boussard, and
  Shah}]{banda2018advances}
Juan~M Banda, Martin Seneviratne, Tina Hernandez-Boussard, and Nigam~H Shah.
  2018.
\newblock Advances in electronic phenotyping: from rule-based definitions to
  machine learning models.
\newblock \emph{Annual review of biomedical data science}, 1:53--68.

\bibitem[{Chevyrev and Kormilitzin(2016)}]{chevyrev2016primer}
Ilya Chevyrev and Andrey Kormilitzin. 2016.
\newblock A primer on the signature method in machine learning.
\newblock \emph{arXiv preprint arXiv:1603.03788}.

\bibitem[{Davenport and Kalakota(2019)}]{davenport2019potential}
Thomas Davenport and Ravi Kalakota. 2019.
\newblock The potential for artificial intelligence in healthcare.
\newblock \emph{Future healthcare journal}, 6(2):94.

\bibitem[{Devlin et~al.(2018)Devlin, Chang, Lee, and
  Toutanova}]{devlin2018bert}
Jacob Devlin, Ming-Wei Chang, Kenton Lee, and Kristina Toutanova. 2018.
\newblock Bert: Pre-training of deep bidirectional transformers for language
  understanding.
\newblock \emph{arXiv preprint arXiv:1810.04805}.

\bibitem[{Gu et~al.(2020)Gu, Tinn, Cheng, Lucas, Usuyama, Liu, Naumann, Gao,
  and Poon}]{gu2020domain}
Yu~Gu, Robert Tinn, Hao Cheng, Michael Lucas, Naoto Usuyama, Xiaodong Liu,
  Tristan Naumann, Jianfeng Gao, and Hoifung Poon. 2020.
\newblock Domain-specific language model pretraining for biomedical natural
  language processing.
\newblock \emph{arXiv preprint arXiv:2007.15779}.

\bibitem[{He et~al.(2016)He, Zhang, Ren, and Sun}]{he2016deep}
Kaiming He, Xiangyu Zhang, Shaoqing Ren, and Jian Sun. 2016.
\newblock Deep residual learning for image recognition.
\newblock In \emph{Proceedings of the IEEE conference on computer vision and
  pattern recognition}, pages 770--778.

\bibitem[{Huang et~al.(2020)Huang, Altosaar, and
  Ranganath}]{huang2019clinicalbert}
Kexin Huang, Jaan Altosaar, and Rajesh Ranganath. 2020.
\newblock Clinicalbert: Modeling clinical notes and predicting hospital
  readmission.
\newblock In \emph{Proc. ACM Conference on Health, Inference, and Learning
  (CHIL)}.

\bibitem[{Johnson et~al.(2016)Johnson, Pollard, Shen, Li-Wei, Feng, Ghassemi,
  Moody, Szolovits, Celi, and Mark}]{johnson2016mimic}
Alistair~EW Johnson, Tom~J Pollard, Lu~Shen, H~Lehman Li-Wei, Mengling Feng,
  Mohammad Ghassemi, Benjamin Moody, Peter Szolovits, Leo~Anthony Celi, and
  Roger~G Mark. 2016.
\newblock Mimic-iii, a freely accessible critical care database.
\newblock \emph{Scientific data}, 3(1):1--9.

\bibitem[{Kalyan and Sangeetha(2020)}]{kalyan2020secnlp}
Katikapalli~Subramanyam Kalyan and S~Sangeetha. 2020.
\newblock Secnlp: A survey of embeddings in clinical natural language
  processing.
\newblock \emph{Journal of biomedical informatics}, 101:103323.

\bibitem[{Kidger et~al.(2019)Kidger, Bonnier, Arribas, Salvi, and
  Lyons}]{kidger2019deep}
Patrick Kidger, Patric Bonnier, Imanol~Perez Arribas, Cristopher Salvi, and
  Terry Lyons. 2019.
\newblock Deep signature transforms.
\newblock In \emph{Advances in Neural Information Processing Systems}, pages
  3105--3115.

\bibitem[{Kidger and Lyons(2020)}]{signatory}
Patrick Kidger and Terry Lyons. 2020.
\newblock \href {https://github.com/patrick-kidger/signatory} {{Signatory:
  differentiable computations of the signature and logsignature transforms, on
  both CPU and GPU}}.
\newblock \emph{arXiv:2001.00706}.

\bibitem[{Lee et~al.(2020)Lee, Yoon, Kim, Kim, Kim, So, and
  Kang}]{lee2020biobert}
Jinhyuk Lee, Wonjin Yoon, Sungdong Kim, Donghyeon Kim, Sunkyu Kim, Chan~Ho So,
  and Jaewoo Kang. 2020.
\newblock Biobert: a pre-trained biomedical language representation model for
  biomedical text mining.
\newblock \emph{Bioinformatics}, 36(4):1234--1240.

\bibitem[{Lyons(2014)}]{lyons2014rough}
Terry Lyons. 2014.
\newblock Rough paths, signatures and the modelling of functions on streams.
\newblock In \emph{Proceedings of the International Congress of
  Mathematicians}, pages 163--184.

\bibitem[{Lyons(1998)}]{lyons1998differential}
Terry~J Lyons. 1998.
\newblock Differential equations driven by rough signals.
\newblock \emph{Revista Matem{\'a}tica Iberoamericana}, 14(2):215--310.

\bibitem[{Morrill et~al.(2020)Morrill, Kormilitzin, Nevado-Holgado,
  Swaminathan, Howison, and Lyons}]{morrill2020utilization}
James~H Morrill, Andrey Kormilitzin, Alejo~J Nevado-Holgado, Sumanth
  Swaminathan, Samuel~D Howison, and Terry~J Lyons. 2020.
\newblock Utilization of the signature method to identify the early onset of
  sepsis from multivariate physiological time series in critical care
  monitoring.
\newblock \emph{Critical Care Medicine}, 48(10):e976--e981.

\bibitem[{Peng et~al.(2019)Peng, Yan, and Lu}]{peng2019transfer}
Yifan Peng, Shankai Yan, and Zhiyong Lu. 2019.
\newblock Transfer learning in biomedical natural language processing: An
  evaluation of bert and elmo on ten benchmarking datasets.
\newblock In \emph{Proceedings of the 18th BioNLP Workshop and Shared Task},
  pages 58--65.

\bibitem[{Peters et~al.(2018)Peters, Neumann, Iyyer, Gardner, Clark, Lee, and
  Zettlemoyer}]{peters2018deep}
Matthew~E Peters, Mark Neumann, Mohit Iyyer, Matt Gardner, Christopher Clark,
  Kenton Lee, and Luke Zettlemoyer. 2018.
\newblock Deep contextualized word representations.
\newblock In \emph{Proceedings of NAACL-HLT}, pages 2227--2237.

\bibitem[{Shaw et~al.(2018)Shaw, Uszkoreit, and Vaswani}]{shaw2018self}
Peter Shaw, Jakob Uszkoreit, and Ashish Vaswani. 2018.
\newblock Self-attention with relative position representations.
\newblock In \emph{Proceedings of the 2018 Conference of the North American
  Chapter of the Association for Computational Linguistics: Human Language
  Technologies, Volume 2 (Short Papers)}, pages 464--468.

\bibitem[{Si et~al.(2019)Si, Wang, Xu, and Roberts}]{si2019enhancing}
Yuqi Si, Jingqi Wang, Hua Xu, and Kirk Roberts. 2019.
\newblock Enhancing clinical concept extraction with contextual embeddings.
\newblock \emph{Journal of the American Medical Informatics Association},
  26(11):1297--1304.

\bibitem[{Toth and Oberhauser(2020)}]{tothbayesian}
Csaba Toth and Harald Oberhauser. 2020.
\newblock Bayesian learning from sequential data using gaussian processes with
  signature covariances.
\newblock In \emph{Proceedings of the International Conference on Machine
  Learning (ICML)}.

\bibitem[{Vaswani et~al.(2017)Vaswani, Shazeer, Parmar, Uszkoreit, Jones,
  Gomez, Kaiser, and Polosukhin}]{vaswani2017attention}
Ashish Vaswani, Noam Shazeer, Niki Parmar, Jakob Uszkoreit, Llion Jones,
  Aidan~N Gomez, {\L}ukasz Kaiser, and Illia Polosukhin. 2017.
\newblock Attention is all you need.
\newblock In \emph{Advances in neural information processing systems}, pages
  5998--6008.

\bibitem[{Wang et~al.(2019)Wang, Liakata, Ni, Lyons, Nevado-Holgado, and
  Saunders}]{wang2019path}
Bo~Wang, Maria Liakata, Hao Ni, Terry Lyons, Alejo~J Nevado-Holgado, and Kate
  Saunders. 2019.
\newblock A path signature approach for speech emotion recognition.
\newblock In \emph{Interspeech 2019}, pages 1661--1665. ISCA.

\bibitem[{Wang et~al.(2020)Wang, Wu, Taylor, Lyons, Liakata, Nevado-Holgado,
  and Saunders}]{wang2020learning}
Bo~Wang, Yue Wu, Niall Taylor, Terry Lyons, Maria Liakata, Alejo~J
  Nevado-Holgado, and Kate~EA Saunders. 2020.
\newblock Learning to detect bipolar disorder and borderline personality
  disorder with language and speech in non-clinical interviews.
\newblock In \emph{Interspeech 2020}. ISCA.

\bibitem[{Wang et~al.(2018)Wang, Wang, Rastegar-Mojarad, Moon, Shen, Afzal,
  Liu, Zeng, Mehrabi, Sohn et~al.}]{wang2018clinical}
Yanshan Wang, Liwei Wang, Majid Rastegar-Mojarad, Sungrim Moon, Feichen Shen,
  Naveed Afzal, Sijia Liu, Yuqun Zeng, Saeed Mehrabi, Sunghwan Sohn, et~al.
  2018.
\newblock Clinical information extraction applications: a literature review.
\newblock \emph{Journal of biomedical informatics}, 77:34--49.

\bibitem[{Xie et~al.(2018)Xie, Sun, Jin, Ni, and Lyons}]{xie2018learning}
Zecheng Xie, Zenghui Sun, Lianwen Jin, Hao Ni, and Terry Lyons. 2018.
\newblock Learning spatial-semantic context with fully convolutional recurrent
  network for online handwritten chinese text recognition.
\newblock \emph{IEEE transactions on pattern analysis and machine
  intelligence}, 40(8):1903--1917.

\bibitem[{Yang et~al.(2016)Yang, Jin, and Liu}]{yang2016deepwriterid}
Weixin Yang, Lianwen Jin, and Manfei Liu. 2016.
\newblock Deepwriterid: An end-to-end online text-independent writer
  identification system.
\newblock \emph{IEEE Intelligent Systems}, 31(2):45--53.

\bibitem[{Yang et~al.(2017)Yang, Lyons, Ni, Schmid, Jin, and
  Chang}]{yang2017leveraging}
Weixin Yang, Terry Lyons, Hao Ni, Cordelia Schmid, Lianwen Jin, and Jiawei
  Chang. 2017.
\newblock Leveraging the path signature for skeleton-based human action
  recognition.
\newblock \emph{arXiv preprint arXiv:1707.03993}.

\end{thebibliography}

\clearpage
\appendix

\section{Appendices}
\label{sec:appendix}

\subsection{Implementation Details}
\label{sec:imp-de}
Our classifier for \textsc{Indication} is a two-layer network, in which the first layer has the size of 50 and 5 for the second layer. ReLU activation and dropout are used in-between the two layers. The dropout rate is set to be 0.1. We also use a two-layer network for the \textsc{Quantity} regression task, in which the first layer has the size of 10 and its second layer predicts the value for \textit{quantity}. We also add a dropout the rate 0.1 in-between the two layers. Our classifier for \textsc{Quantity Tag} has only one layer the size of 5. 

We reduce the dimension of attended vector before signature transform from 768 to 32, as depicted in Figure~\ref{fig:sig-att} (left). We set the order of truncated signature to be 2, and employ 8 parallel attention layers (or heads). We ran grid-search over two choices of learning rate $lr: \left [  0.00003, 0.00005 \right ]$, and 10 randomly sampled triplets of $\alpha_{qnt}$, $\beta_{qntt}$ and $\beta_{ind}$. Each value of $\alpha_{qnt}$, $\beta_{qntt}$ and $\beta_{ind}$ is between $0$ and $1$ respectively, while we make sure each combination of the three values sums to $1$.

\subsection{Signature of paths}
\label{sec:bg-sig}
We begin with the definition of the signature, using more traditional notation of stochastic calculus.
\begin{definition}
Let $X = (X^1,...,X^d)$ be a path in $\mathbb{R}^d$. The signature of $X$ is defined as the infinite collection of iterated integrals:
\begin{equation*}
\begin{aligned}
S(X) &= \left({\int ... \int}_{a < t_1 < ... < t_k < b} dX_{t_1} \otimes ... \otimes dX_{t_k}\right)_{k \geq 0} \\
& = \left(\left({\int ... \int}_{a < t_1 < ... < t_k < b} dX_{t_1}^{i_1} ...  dX_{t_k}^{i_k}\right)_{1 \leq i_1,...,i_k \leq d} \right)_{k \geq 0}
\end{aligned}
\end{equation*}
\noindent where $dX_t = \frac{dX_t}{dt}dt$ and the $k=0$ term is taken to be 1 $\in \mathbb{R}$.
\end{definition}

\begin{definition}
The truncated signature of order $N$ of $X$ is defined as:
\begin{equation*}
\begin{aligned}
S^{N}(X) &= \left({\int ... \int}_{a < t_1 < ... < t_k < b} dX_{t_1} \otimes ... \otimes dX_{t_k}\right)_{0 \leq k \leq N}.
\end{aligned}
\end{equation*}
\end{definition}

\noindent The dimension of the truncated signature explodes exponentially with the input path dimension: 

\begin{proposition} \label{nbcomponents}
For any $d \geq 1$, the truncated signature of order $N$ of a \textit{d}-dimensional path has the dimension of:
\begin{equation*}
\begin{aligned}
\sum_{k=0}^{N} d^k = \frac{d^{N+1} - 1}{d-1}
\end{aligned}
\end{equation*}
\end{proposition}

\noindent In practice, the term of order 0 is dropped as it is always equal to 1. For clarity, we define:
\begin{equation*}
\begin{aligned}
S(X)^{i_1,...,i_k} =  {\int ... \int}_{a < t_1 < ... < t_k < b} dX_{t_1}^{i_1} ...  dX_{t_k}^{i_k}
\end{aligned}
\end{equation*} 
\noindent with $1 \leq i_1,...,i_k \leq d$, so that: 

\begin{equation*}
\begin{aligned}
    S(X) &= \left(\left(S(X)^{i_1,...,i_k})_{1 \leq i_1,...,i_k \leq d}\right)\right)_{k \geq 0} \\
    &= (1, S(X)^1,...,S(X)^d,S(X)^{1,1},S(X)^{1,2},...).
\end{aligned}
\end{equation*}

\begin{table*}[htb]
\begin{center}
\begin{tabular}{| *{3}{c|} }
\hline
$d_{presig}$ & $\mathrm{order}_{sig}$ & $d_{sig}$ \\
    \hline
 512  &   1  &  512  \\
 512  &   2  &  262K  \\
 256  &   2  &  66K  \\
 128  &   2  &  16K  \\
 128  &   3  &  2M  \\
 64   &   2  &  4K  \\
 64   &   3  &  266K \\
 32   &   2  &  1057  \\
 32   &   3  &  34K   \\
   16   &   2  &  272   \\
   \hline
\end{tabular}
\quad
\begin{tabular}{| *{3}{c|} }
\hline
$d_{presig}$ & $\mathrm{order}_{sig}$ & $d_{sig}$ \\
    \hline
 16   &   3  &  4K  \\
  8   &   2  &  72  \\
  8   &   3  &  584 \\
 8   &   4  &   5K  \\
 4   &   4  &  340   \\
 4   &   5  &  1365   \\
 4   &   6  &  5K   \\
  2   &   9  &  1022  \\
 2   &   10  &  2K  \\
 2   &   12  &  8K  \\   
   \hline
\end{tabular}
\end{center}
\caption{The number of dimensions ($d_{sig}$) of the truncated signature is determined by the size of its input ($d_{presig}$) and the order of truncation selected ($\mathrm{order}_{sig}$).}
\label{tab:dims}
\end{table*}

\begin{figure*}[htb]
 \begin{center}
  \includegraphics[width=1.0\textwidth]{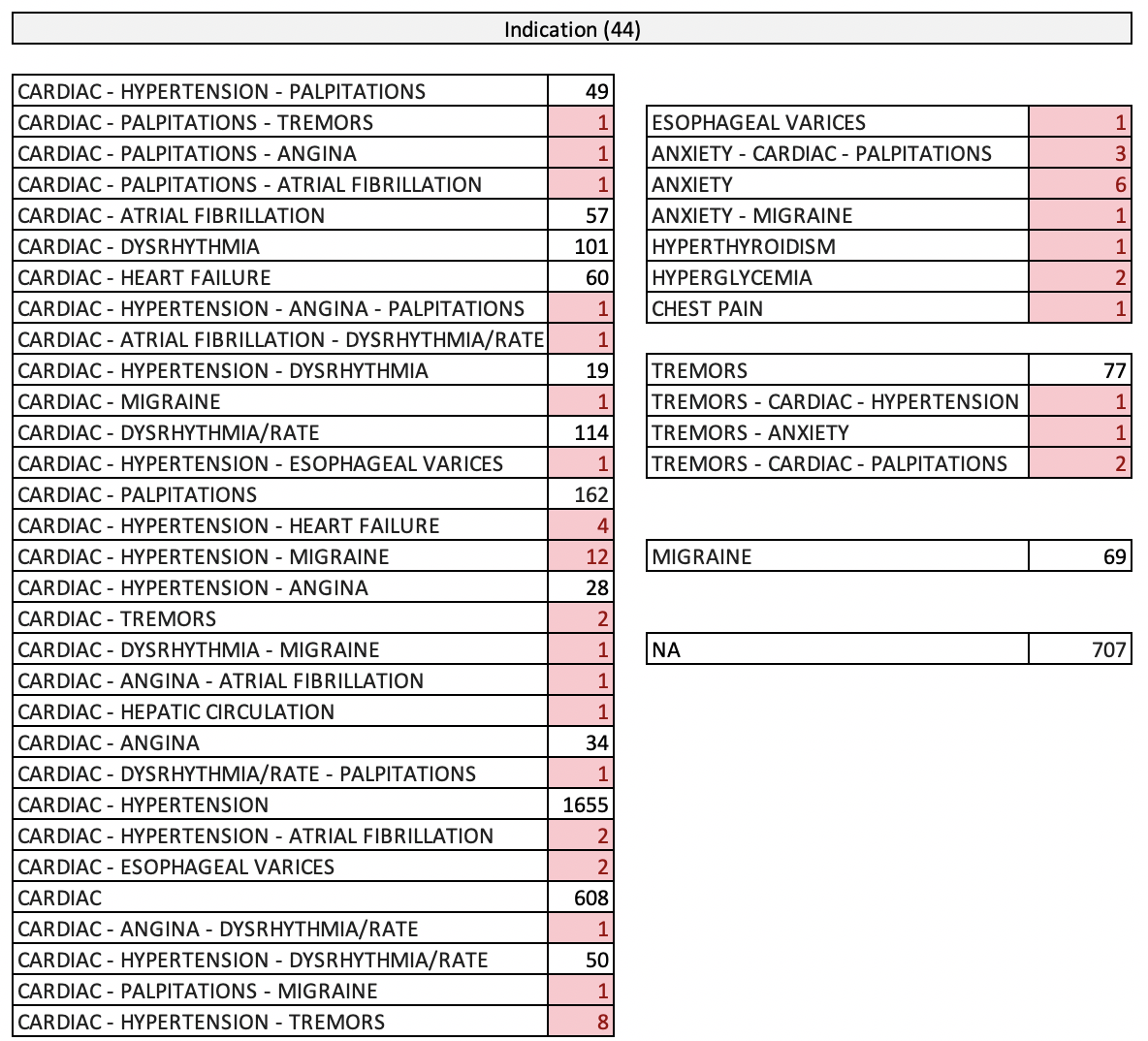}
  \caption{Number of prescriptions per indication class, where the indication label has the original 44 classes. \textit{NA} stands for \textit{Not Annotated}.}
  \label{fig:indication-original}
 \end{center}
\end{figure*}

\begin{table*}[htb]
\begin{center}
\scalebox{1.0}{
\begin{tabular}{|c|ccccc|c|}
\hline
 \multicolumn{1}{|c|}{\multirow{2}{*}{\textbf{Model}} } &
 \multicolumn{5}{|c|}{\textbf{\textsc{Quantity Tag}}} &
 \multicolumn{1}{|c|}{\multirow{2}{*}{\textbf{\textsc{Quantity}}} } \\
 
 \multicolumn{1}{|c|}{} &
 \multicolumn{1}{|c|}{\textbf{Standard}} &
 \multicolumn{1}{|c|}{\textbf{APPP}} &
 \multicolumn{1}{|c|}{\textbf{PRN}} &
 \multicolumn{1}{|c|}{\textbf{Complex}} &
 \multicolumn{1}{|c|}{\textbf{NS}} &
 \multicolumn{1}{|c|}{} \\
\hline
Base & 0.71 & 0.50 & 0.10 & 0.76 & 0.95 & 0.50 \\
ClinicalBERT & 0.83 & 0.97 & 0.89 & 0.99 & 0.79 & 0.21 \\
M-BERT & 0.94 & 0.41 & 0.04 & 0.81 & 0.98 & 0.41 \\
\hline
Base + LSTM & 0.93 & 0.36 & 0.99 & 0.77 & 0.74 & 0.45 \\
ClinicalBERT + LSTM & 0.90 & 0.23 & 0.99 & 0.61 & 0.49 & 0.47 \\
M-BERT + LSTM & 0.29 & 0.99 & 0.72 & 0.47 & 0.93 & 0.50 \\
\hline
Base + STE & 0.97 & 0.84 & 0.77 & 0.44 & 0.88 & 0.23 \\
ClinicalBERT + STE & 0.99 & 0.70 & 0.85 & 0.65 & 0.65 & 0.36 \\
M-BERT + STE & 0.86 & 0.97 & 1.00 & 0.89 & 0.86 & 0.15 \\
\hline
\end{tabular}}
\end{center}
\caption{Model performance comparison for \textsc{Quantity} and also across different classes in \textsc{Quantity Tag}. APPP: As Per Previous Prescription; NS: Not Specified.}
\label{tab:res-xl-quant}
\end{table*}

\begin{table*}[htb]
\begin{center}
\scalebox{1.0}{
\begin{tabular}{|c|ccccc|}
\hline
 \multicolumn{1}{|c|}{\multirow{2}{*}{\textbf{Model}} } &
 \multicolumn{5}{|c|}{\textbf{\textsc{Indication}}} \\
 
 \multicolumn{1}{|c|}{} &
 \multicolumn{1}{|c|}{\textbf{Cardiac}} &
 \multicolumn{1}{|c|}{\textbf{Tremors}} &
 \multicolumn{1}{|c|}{\textbf{Migraine}} &
 \multicolumn{1}{|c|}{\textbf{Others}} &
 \multicolumn{1}{|c|}{\textbf{NA}} \\
\hline
Base & 0.03 & 0.05 & 0.02 & 0.27 & 0.03 \\
ClinicalBERT & 0.00 & 0.24 & 0.03 & 0.04 & 0.13 \\
M-BERT & 0.05 & 0.01 & 0.09 & 0.35 & 0.00 \\
\hline
Base + LSTM & 0.02 & 0.00 & 0.00 & 0.31 & 0.00 \\
ClinicalBERT + LSTM & 0.01 & 0.03 & 0.34 & 0.08 & 0.02 \\
M-BERT + LSTM & 0.05 & 0.00 & 0.00 & 0.04 & 0.03 \\
\hline
Base + STE & 0.00 & 0.00 & 0.02 & 0.08 & 0.05 \\
ClinicalBERT + STE & 0.00 & 0.26 & 0.05 & 0.02 & 0.00 \\
M-BERT + STE & 0.00 & 0.11 & 0.00 & 0.00 & 0.15 \\
\hline
\end{tabular}}
\end{center}
\caption{Model performance comparison across different classes in \textsc{Indication}. NA: Not Annotated.}
\label{tab:res-xl-ind}
\end{table*}

\begin{sidewaystable*}[htb]
\begin{center}
\scalebox{0.95}{
\begin{tabular}{|c|c|c|c|c|}
\hline
\textbf{Swedish} & \textbf{Translated English} & \textbf{Indication} & \textbf{Quantity} & \textbf{Quantity Tag}  \\
\hline
\specialcell{0.5 TABLETTER 2 GANGER DAGLIGEN.\\ FOR HJARTRYTMEN OCH SANKER BLODTRYCKET\\ TABLETTERNA SVALJES HELA\\ (KAN DELAS VID SVALJSVARIGHETER,\\ MEN FAR EJ K} & \specialcell{0.5 tablets 2 times daily.\\ For the heart rhythm and reduces blood pressure\\ The tablets are swallowed whole \\ (can be divided at swallowing durations,\\ but don't k} & \specialcell{Cardiac-\\hypertension} & 1 & Standard\\
 \hline
\specialcell{1 TABLETT 1 GANG DAGLIGEN I 3 VECKOR. \\EVENTUELL HOJNING TILL 2 TABLETTER DAGLIGEN\\ BEROENDE PA KONTROLLEN OCH SYMTOMLINDRING,\\ MOT HJARTKLAPPNING} & \specialcell{1 table 1 time daily for 3 weeks. \\Successful increase for 2 tablets daily\\ dependent on control and symptoms relief,\\ against palpitations} & \specialcell{Cardiac-\\palpitations} & 1 & Complex  \\
 \hline
\specialcell{1 TABLETT 1 GANG DAGLIGEN.\\ EFTERHAND EVENTUELL OKNING TILL EN TABLETT\\ MORGON OCH LUNCH MEN BORJA MED EN TABLETT\\ TIDIG MORGON, MOT SKAKNINGAR.} & \specialcell{1 tbale 1 time daily.\\ Previously opening to a tablet\\ morning and lunch but begin with a tablet\\ early morning, against shakes} & \specialcell{NA} & 1 & Standard  \\
 \hline
\specialcell{1/2 TABLETT PA MORGONEN , \\1/2 TABLETT PA LUNCHEN OCH 1/2 TABLETT PA KVALLEN\\ MOT TREMOR, BLODTRYCKSREGLERANDE} & \specialcell{1/2 table on morning,\\ 1/2 table on lunch and 1/2 table on evening\\ against tremor, blood pressure control} & \specialcell{NA} & 1.5 & Standard \\
 \hline
\specialcell{1 TABLETT 1 GANG DAGLIGEN MOT HOGT BLODTRYCK\\ DU BOR BESTALLA LAKARTIDLAMPLIGEN MAJ JUNI\\ FOR KONTROLL BT DIABETES.} & \specialcell{1 tablet 1 time daily against high blood pressure\\ you should order the painting lighting may june\\ for control BT diabetes} & \specialcell{Cardiac-\\hypertension} & 1 & Standard  \\
\hline
\specialcell{2 depottabletter kl. 08, 1 depottablett kl. 20. Dagligen.\\ Mot hOgt blodtryck} & \specialcell{2 prolonged-release tablets at. 08,\\ 1 prolonged-release tablet at. 20. Daily.\\ Against high blood pressure} & \specialcell{Cardiac-\\hypertension} & 3 & Standard \\
 \hline
\end{tabular}}
\end{center}
\caption{Longer example prescriptions with translations and annotations.}
\label{tab:data-example-long}
\end{sidewaystable*}

\end{document}